\documentclass[11pt,authoryear]{elsarticle}
\usepackage{graphicx}
\usepackage[colorlinks=true,linkcolor=black, citecolor=blue, urlcolor=blue]{hyperref}
\usepackage{lineno}
\usepackage[utf8]{inputenc}
\usepackage{arabtex}
\usepackage{utf8}
\setcode{utf8}
\usepackage{color,soul}
\usepackage[paperheight=11in,paperwidth=9in,left=25mm,right=25mm,top=25mm,bottom=25mm]{geometry}

\usepackage{natbib}
\usepackage[labelsep=period]{caption}
\usepackage[labelfont=bf]{caption}
\immediate\write18{cp `kpsewhich agsm.bst` .}
\def\endabstract{\egroup}
\journal{Journal of King Saud University - Computer and Information Sciences}
\begin{document}
\begin{frontmatter}
\makeatletter
\def\ps@pprintTitle{%
 \let\@oddhead\@empty
 \let\@evenhead\@empty
 \def\@oddfoot{}%
 \let\@evenfoot\@oddfoot}
\makeatother
\title{Arabic aspect based sentiment analysis using bidirectional GRU based models}



\author[add1]{Mohammed M.Abdelgwad}
\ead{mohammed.mustafa@aun.edu.eg}
\author[add1]{Taysir Hassan A Soliman}
\ead{taysirhs@aun.edu.eg}
\author[add1]{Ahmed I.Taloba}
\ead{Taloba@aun.edu.eg}
\author[add1]{Mohamed Fawzy Farghaly}
\ead{mfseddik@aun.edu.eg}

\address[add1]{
Information system department\\*
     Faculty of computer and information \\*
     Assiut university\\*
     Egypt}

\begin{abstract}
Aspect-based Sentiment analysis (ABSA) accomplishes a fine-grained analysis that defines the aspects of a given document or sentence and the sentiments conveyed regarding each aspect. This level of analysis is the most detailed version that is capable of exploring the nuanced viewpoints of the reviews.
The bulk of study in ABSA focuses on English with very little work available in Arabic.
Most previous work in Arabic has been based on regular methods of machine learning that mainly depends on a group of rare resources and tools for analyzing and processing Arabic content such as lexicons, but the lack of those resources presents another challenge. In order to address these challenges, Deep Learning (DL)-based methods are proposed using two models based on Gated Recurrent Units (GRU) neural networks for ABSA. The first is a DL model that takes advantage of word and character representations by combining bidirectional GRU, Convolutional Neural Network (CNN), and Conditional Random Field (CRF) making up the (BGRU-CNN-CRF) model to extract the main opinionated aspects (OTE). The second is an interactive attention network based on bidirectional GRU (IAN-BGRU) to identify sentiment polarity toward extracted aspects. We evaluated our models using the benchmarked Arabic hotel reviews dataset proposed by \citep{pontiki-etal-2016-semeval}. The results indicate that the proposed methods are better than baseline research on both tasks having 39.7\% enhancement in F1-score for opinion target extraction (T2) and 7.58\% in accuracy for aspect-based sentiment polarity classification (T3). Achieving F1 score of 70.67\% for T2, and accuracy of 83.98\% for T3.
\end{abstract}

\begin{keyword}
Aspect-based sentiment analysis (ABSA) \sep deep learning \sep opinion target extraction (OTE) \sep aspect sentiment polarity classification \sep BGRU-CNN-CRF and IAN-BGRU.
\end{keyword}
\end{frontmatter}

\section{Introduction}
Advances in web technology have created new opportunities to interact with user-generated content, such as blogs, social networks, forums, website reviews, etc. \citep{rana2016aspect}. With this remarkable increase in data and the complexity of managing unstructured texts in natural languages, individuals and organizations in the field of data mining have been very interested in taking advantage of this flow.\\
Sentiment Analysis (opinion mining or emotional analysis) is a computer-generated examination of human opinions, moods, and emotions \citep{zhao2016sentiment}. Aiming to determine the attitude of the author of a specific piece of content related to the subject of interest. It determines the polarity of the document (review, tweet, or news) whether the opinion expressed is positive, negative, or neutral. \\
There are three approaches for Arabic sentiment analysis: corpus-based, lexicon-based, and hybrid-based(the combination of lexicon and corpus) \citep{al2018deep}. The first publication on Arabic SA was in 2006 as indicated in \citep{al2018deep}. Since that time, several methods have been published in Arabic SA.\\
Studying sentiment analysis is possible at three levels as argued by \citep{hu2004mining}: the document level assumes that there is only one opinion in the entire document, so the task will be to determine the positive, negative, or neutral opinion for the entire document. The task at sentence level breaks down the document into a group of sentences and decides whether a positive, negative or neutral opinion should be expressed in each sentence. Neither the document level nor the sentence level analysis could determine precisely what people liked and disliked. As sentences or reviews may include many opinions on different aspects of a particular entity. In addition, these opinions may conflict with one another. These reviews require another suitable SA called ABSA.\\
Three main ABSA tasks can be identified as proposed by \citep{pontiki-etal-2016-semeval}; T1: aspect category identification, T2: aspect opinion target extraction, and T3: aspect polarity detection. The topics of this study are T2 and T3 tasks.\\
According to \citep{al2018deep} there are several differences between SA and ABSA; such as (a) linking text parts to specific aspects (i.e. extracting target opinion expressions), and (b) Paraphrasing the text by extracting text parts that discuss the same aspects (i.e. Battery efficiency and power usage both related to the same aspect).\\
ABSA has been a major focus of high-profile Natural Language Processing (NLP) conferences and workshops like SemEval due to its importance. SemEval is an annual NLP workshop that offers a number of activities to the scientific community to test SA systems. The first ABSA joint task was coordinated by SemEval in 2014 \citep{pontiki-etal-2014-semeval}. This task provided the scientific community with both standard datasets and joint evaluation procedures.
ABSA activities were effectively replicated over the next two years at SemEval \citep{pontiki-etal-2015-semeval},\citep{pontiki-etal-2016-semeval} as the task expanded to include different domains, languages, and problems. In fact, SemEval-2016 presented a total of 39 datasets for the ABSA task in 7 domains and 8 languages. In addition, a classifier proven to perform well for NLP tasks, which is Support Vector Machine (SVM), was used in the baseline evaluation procedure.\\
More recently, experimental work with innovative machine learning methods, called “Deep Learning” multi-layer processing technology that utilizes consecutive unit layers to build on previous outputs, was demonstrated using the backpropagation algorithm \citep{lecun2015deep}. On each layer, the inputs are converted to numerical representations, which are later classified. Therefore, an increasingly greater degree of abstraction is obtained \citep{goodfellow2016deep}.\\
DL is considered as one of the highly suggested techniques in machine learning for various NLP challenges such as SA \citep{kwaik2019lstm,luo2019network}, machine translation \citep{ameur2017arabic,li2019implementing}, named entity recognition \citep{khalifa2019character}, and speech recognition \citep{zerari2019bidirectional,algihab2019arabic}. The strength of DL is that, aside from its great performance, it does not rely on handcrafted features or external resources.\\
Word embedding or distributed representations improve neural network performance and enhances DL models. Two common ways to embed words available in Arabic: Word2Vec \citep{mikolov2013efficient} and FastText \citep{bojanowski2017enriching}.\\
First, Word2Vec makes use of small neural networks to calculate word embedding based on word context. There are two ways to put that approach into practice. The first is a continuous bag of words or CBOW. In this method, the network attempts to predict which word is most likely given the context. Skip-gram is the second approach, the idea is very similar but the network is working in the opposite direction, the network attempts to predict the context given target word. In several areas of NLP, Word2Vec has proved useful. But one unresolved issue has been unknown term generalization. Second, FastText which was founded in 2016 by Facebook vowed to resolve this obstacle.\\
A new development in DL has emerged, it is the attention mechanism. The attention mechanism has achieved good success in computer vision and in many NLP applications such as document sentiment classification, document summarization, named entity recognition, and machine translation. The attention mechanism of a neural network allows learning properly by focusing selectively on the major parts of the sentence while performing a task.\\
Recently, the attention mechanism has been relied upon in several DL-based models for SA \citep{ma2017interactive,huang2018aspect,liu2018content,yang2018feature}. The attention mechanism allows the neural network to concentrate on the various parts of a sentence that relate to each aspect when the sentence contains different aspects. \\
In several NLP tasks, Interactive attention networks(IAN) has shown impressive results in machine translation \citep{meng2016interactive}, question answering \citep{wang2016inner,li2017context}, and document classification \citep{yang2016hierarchical}. Authors of \citep{ma2017interactive} suggested using IAN for English ABSA and achieved competitive results. The main idea is interactive learning to represent targets and contexts. To further improve the representation of targets and context, we proposed using the features provided by the GRU model in general and bidirectional GRU in particular instead of using single direction Long Short Term Memory (LSTM) in the base model, the bi-directional GRU overcomes feed-forward models limited ability by extracting unlimited contextual information from both sentence directions.\\

\begin{figure}[!ht]
\begin{center}
 \includegraphics[width=.7\textwidth]{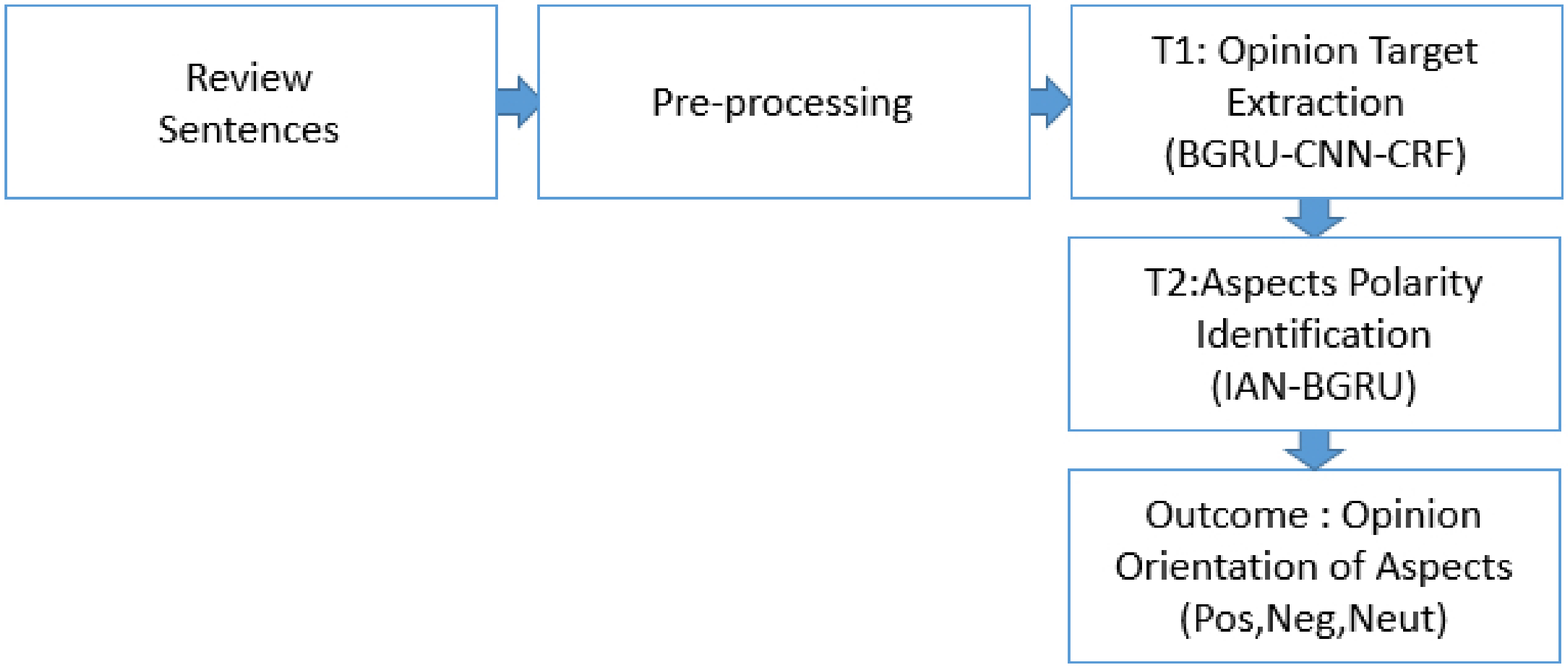}
    \caption{The overall workflow for the proposed ABSA approach.}
    \label{fig:Figure1}
\end{center}
\end{figure}

To carry out the ABSA tasks, we follow some steps that ultimately make up the ABSA workflow, and they are in order:
\begin{enumerate}
  \item 
  Breaking down reviews into individual sentences.
  \item 
  preprocessing each sentence (tokenization, stopwords removal, and text vectorization).
  \item
  T1: Extracting main opinionated aspects using (BGRU-CNN-CRF) model.
  \item 
  T2: Determine sentiments related to each aspect using (IAN-BGRU) model.
\end{enumerate}
The overall workflow for the proposed ABSA approach is shown in \autoref{fig:Figure1}.

In this paper, ABSA research tasks are performed by a type of RNN (GRU), where two GRU models were built, as follows: 
(a) DL architecture based on a state-of-the-art model that utilizes the representations of both words and characters through the combination of bidirectional GRU, CNN, and CRF (BGRU-CNN-CRF) to extract the main opinionated aspects (i.e. T2: (OTE))
(b) IAN based on bidirectional GRU (IAN-BGRU) is implemented to identify sentiment polarity toward extracted aspects from T2 (i.e. T3).\\
The main contributions of this study are :

\begin{enumerate}
  \item 
  The proposed models did not rely on any handcrafted features or external resources such as lexicons, which considered one of the tools that are not widely available in the public domain for analyzing and processing Arabic content and requires great effort in collection.
  \item 
  The proposed models are better than baseline research on both tasks having 39.7\% enhancement in F1-score for opinion target extraction (T2) and 7.58\% in accuracy for aspect-based sentiment polarity classification (T3). Obtaining F1 score of 70.67\% for T2, and accuracy of 83.98\% for T3.
\end{enumerate}

The rest of this paper is arranged as follows: Section 2 addresses literature reviews of ABSA; Section 3 illustrates the proposed models; Section 4 explains the dataset and the baseline approach; Section 5 Presents results and discussion; section 6 discuss case study; finally, section 7 concludes the paper and outlines future work plans. 

\section{Related work}
This section is divided into two main sub-sections covering related works in English and Arabic ABSA separately, each subsection highlights the effective studies that have been applied on OTE and aspect sentiment polarity tasks.
\subsection{English ABSA}
\subsubsection{Opinion Target Extraction (OTE)}
Deep learning methods were used for the first time on the Opinion target extraction task by \citep{irsoy2014opinion} instead of using regular CRF model. They could extract opinion targets using deep recurrent neural networks.\\
The authors of \citep{liu2015fine} also applied different RNN variations on this task like LSTM supported with word embeddings and some handcrafted features and achieved better results than CRF models. \\
The authors of \citep{da2019aspect} applied CNN model supported with word embeddings and POS embeddings to make sequence labeling easier. Two types of word embeddings were proposed: general-purpose and domain-specific word embeddings. The best results were achieved when applying the model with POS and domain-specific word embeddings.\\
\citep{chen2017improving} applied Bi-LSTM with CRF on top, to extract opinion aspects. They experimented with many datasets and achieved competitive results.

\subsubsection{aspect sentiment polarity classification}
ABSA is a SA branch where research approaches can be categorized into two approaches: regular methods of machine learning and methods based on DL. Sentiment classification at aspect level is usually considered to be a problem of text classification.\\
Text classification methods, like SVM can be applied to ABSA without taking into consideration the specified targets, as indicated in \citep{pang2002thumbs}. \\
Several rule-based methods have been developed to deal with ABSA in some early works, as indicated in \citep{nasukawa2003sentiment,ding2007utility} where sentence dependency parsing was executed, then predefined rules were used to determine sentiments related to each aspect. Although these methods have achieved satisfactory results, their results rely heavily on the effectiveness of labor-intensive handcraft features.\\
NN variations encourage research in different areas of NLP especially those that need fine-grained analysis like ABSA, as they are capable of generating new representations using original features through several hidden layers.\\
Recursive NN (Rec-NN) can be used to execute semantic compositions on tree structures, so it was adopted to classify sentiments at aspect level by the authors \citep{dong2014adaptive,nguyen2015phrasernn}, by transferring the opinion targets into the root of a tree and propagates the sentiment of targets according to context and syntax relations.\\
RNN, are widely used in ABSA to identify sentiments polarity at aspect level. LSTM is an effective RNN network that is capable of reducing vanishing gradient problems. However, LSTMs are not suitable for addressing the interactive correlation between context and aspect, leading to an enormous loss of aspect relevant data. To include aspects into the model, the authors of \citep{tang2015effective} proposed Target-Dependent LSTM (TD-LSTM) and Target-Connection LSTM (TC-LSTM). The TD-LSTM splits the sentence into the left and right parts around the aspect and flows into two LSTM models in different forward and backward sequential paths. In order to determine the sentiment polarity label, the final hidden vectors of left LSTM and right LSTM are linked to be fed into Softmax layer. Nonetheless, the interactions between aspect target and context are not captured by TD-LSTM. To solve this issue, TC-LSTM uses the semantical interaction between the aspect and the context, by integrating the aspect target and context word embeddings as the inputs, and transfers it back and forth through two different LSTMs, similar to those used in TD-LSTM.\\
The attention mechanism allows the neural network to learn properly by focusing selectively on the major parts of the sentence while performing a task. Attention-based LSTM (ATAE-LSTM) was developed by the authors \citep{wang2016attention} to explore the correlation between aspects and contexts through applying the attention mechanism to assist in identifying the important parts of a sentence towards the stated aspect.
ATAE-LSTM combines embeddings of both the context and aspect and use it as input to LSTM in order to achieve the best possible use of target information. The hidden LSTM vectors would provide knowledge of the aspect target, that may allow the model to obtain attention weights more accurately.\\
Only one attention can fail to capture various key context words associated with different targets at very long distance of dependency, so the authors of \citep{chen2017recurrent} suggested using multiple attentions to deal with this problem by generating recurrent attention on memory (RAM). RAM produces memory from input, and by applying various attentions on memory, it could extract essential information, and for prediction, it uses a non-linear combination of extracted features from different attentions using GRU.\\
The authors of \citep{tang2016aspect} proposed deep memory neural network (MemNet), which consists of applying  multi-hop attention layers on context word embeddings of the sentence and considered the last hop output as the final target representation.\\ 
Review typically consists of many sentences and each sentence consists of several words, so the review structure is hierarchical. Based on the hierarchical structure of the review, the authors of \citep{ruder2016hierarchical} developed Bidirectional Hierarichal LSTM (H-LSTM) for ABSA. They noticed that modeling knowledge in the internal review structure can enhance the performance of the model.\\
To deal with complex sentence structures that have many aspects, authors of \citep{liu2018content} have proposed the CABAC model. Two types of attention were used, one at the sentence level, to focus on words that are important with respect to the aspect, and the other attention type was used to take into account the order and correlation of words using a group of memories.
While effective, LSTM cannot train in parallel and tends to be time consuming as is the case with other RNNs, since they are time-series NN. Therefore a simpler, more accurate, and faster model based on CNN and gating mechanism was proposed by the authors of \citep{xue2018aspect} than conventional LSTM models with attention mechanisms, as their model calculations can easily be paralyzed during training and have no dependence. In their research, they concentrated on only two tasks: aspect category analysis and aspect sentiment analysis, and in both of them, they got great results.\\
CNN also serves as a supplementary method for finding key local features like linguistic patterns (CNN-LP) \citep{poria2016aspect} or as an efficient way to substitute attention for ABSA like Target-Specific Transformation Networks \citep{li2018transformation}.\\
To deal with this task \citep{li2019exploiting} explored BERT embeddings \citep{devlin2018bert} with various simple neural networks such as Linear, GRU, conditional random field, and self-attention layers. The experimental results showed BERT-based neural networks achieved higher results compared to non-BERT complex models.

\subsection{Arabic ABSA}
Despite the large numbers of Arabic speakers and being a rich morphological language, the number of works currently available in the Arabic ABSA still limited, most of which are listed in the next subsection.
\subsubsection{OTE and aspect sentiment polarity classification}
For Arabic ABSA, a few attempts were made, where the HAAD dataset was presented in 2015 along with a baseline study \citep{al2015human}. HAAD was annotated via the SemEval-2014 framework.\\
In support of ABSA task, another benchmarked dataset of Arabic hotel reviews was noted in 2016, this dataset was used to validate some of the methods proposed in the Multilingual ABSA SemEval-2016 Task5.\\
The authors of \citep{al2019enhancing} suggested applying a set of supervised machine learning-based classifiers enhanced with a set of hand-crafted features such as morphological, syntactic, and semantic features on the Arabic hotel reviews dataset.
Their approach covered three tasks: identify aspect categories, extract opinion targets, and identify the sentiment polarity. The evaluation results showed that their approach was very competitive and effective.\\
In addition, the authors of \citep{al2018deep} proposed applying two supervised machine learning-based approaches namely SVM and RNN on the Arabic hotel reviews dataset in line with task 5 of SemEval-2016.
The researchers investigated the three tasks : aspect category identification, (OTE), and sentiment polarity identification. The findings indicate that SVM outperforms RNN in all tasks. Though, the deep RNN was found to be faster and better by comparing the time taken during training and testing.\\
Authors of \citep{al2019using} proposed applying two models based on LSTM and were assessed using Arabic hotel reviews dataset, namely: (a) combination of Bi-directional LSTM and Conditional Random Field classifier (BLSTM- CRF) based on character level and word level for extracting the main opinionated aspects from the text, (b) aspect-based LSTM for sentiment polarity classification (AB-LSTM-PC) used for handling the third task. The test results have demonstrated that their methods surpass the baseline and are pretty effective.\\
last, attention-based neural network was proposed by \citep{al2020extracting} to extract opinion targets. Their model composed of Bidirectional lstm with CNN as encoder and Bidirectional lstm as a decoder with attention and CRF. Their model was applied on the Arabic hotel reviews dataset and achieved state-of-the-art results.\\
In this paper, we investigated the modeling power of BGRU in different Arabic ABSA tasks. For OTE, our model (BGRU-CNN-CRF) depends mainly on BGRU to extract word-level features. Also, two different types of word embeddings were used to train our model (fastText and Word2vec). High results were observed after applying fastText embeddings.
Despite the simplicity of our model, it outperformed the baseline and achieved high results close to \citep{al2020extracting}.\\
For the third task, IAN based on Bidirectional GRU (IAN-BGRU) is implemented to identify sentiment polarity toward extracted aspects. The evaluation results showed that our model is competitive and achieved the state of the art results.

\section{Proposed methods}
In this section, the proposed models previously mentioned in ABSA workflow in Figure \ref{fig:Figure1} will be explained in details. The proposed models are mainly based on GRU, which is a form of RNN. RNN is an Artificial Neural Network (ANN), that is used in different NLP applications. RNN was configured to identify the sequential data properties and then predict the following scenarios using patterns. The main advantage of RNN over feed-forward neural networks is its ability to process inputs of any length and remember all information all the time, which is very useful in any time-series prediction. RNN uses recurrent hidden units, and their activation is based on the previous step each time. The key disadvantages of RNNs are the problems of gradient vanishing/exploding, which make it harder to train and deal with major problems of machine learning \citep{bengio1994learning,pascanu2013difficulty}. GRU has been suggested as a solution to this problem and has proven to be effective in many NLP problems.\\
Accordingly, two models based on GRU have been proposed to handle the research tasks (BGRU-CNN-CRF for T2, and IAN-BGRU for T3).

\subsection{Background}
This part provides a detailed description of all the components used to build our models. They include GRU, BGRU, CNN, and CRF.

\subsubsection{Gated Recurrent Unit (GRU)}

\begin{figure}[!ht]
\begin{center}
 \includegraphics[width=.4\textwidth]{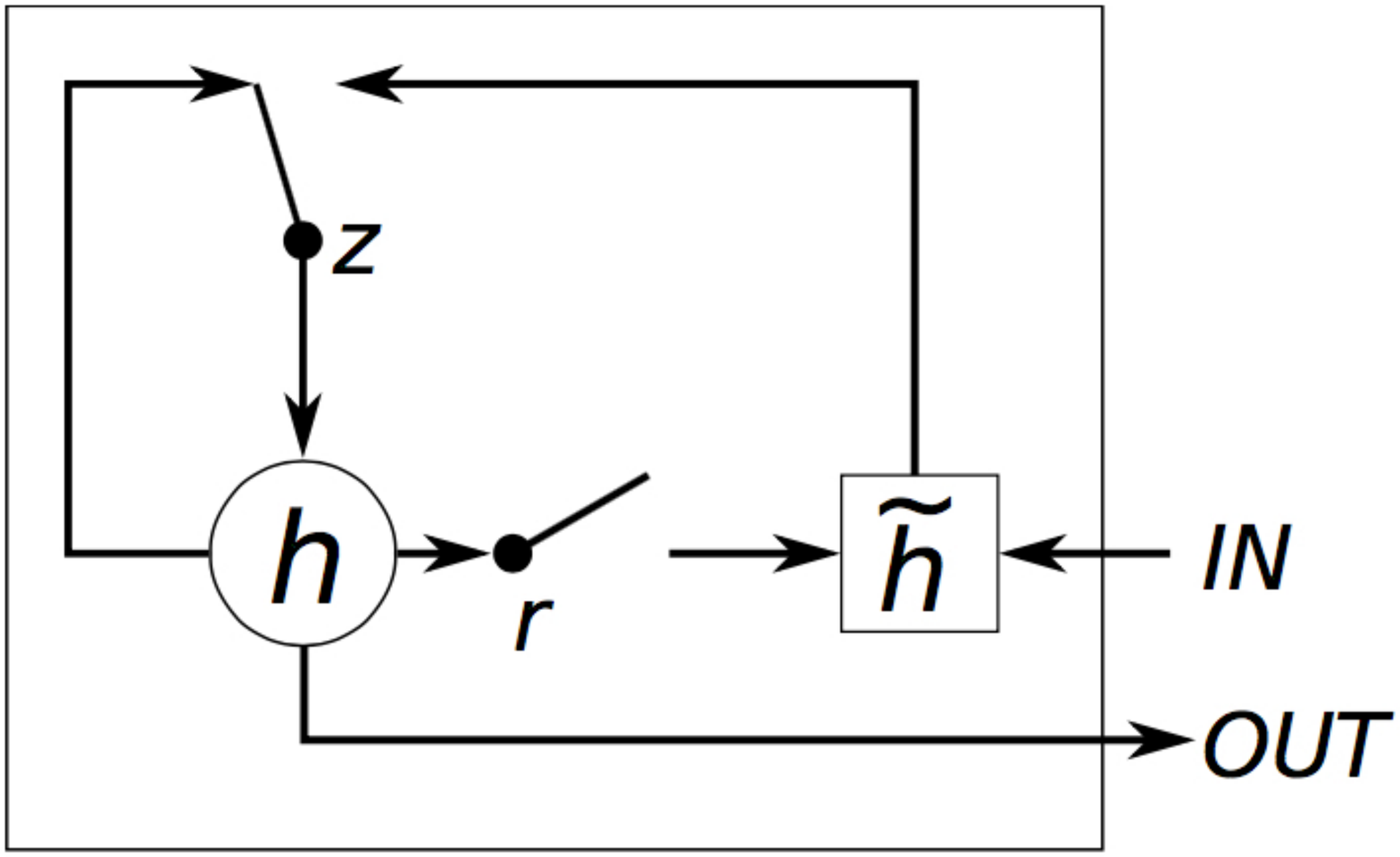}
    \caption{Gated Recurrent Unit.}
    \label{fig:Fig2}
\end{center}
\end{figure}

Recently, GRU \citep{cho2014learning}, a family of RNNs, has been proposed to deal with gradient vanishing/exploding problems. GRU is a powerful and simple alternative to LSTM networks \citep{hochreiter1997long}. Similar to LSTM models, GRU is designed to adaptively update or reset the memory content using \(r^j\) reset gate and a \(z^j\) update gate that are similar to the forget and input gates of LSTM. Compared to LSTM, GRU does not have a memory cell and only has two gates.
The GRU activation \(h_t^j\) in time t is the linear interpolation of previous activation \(h_{t-1}^j\) and candidate activation \(\tilde{h}_t^j\). \\
To compute the state \(h_t^j\) of the j-th GRU at time step t, we use the following equation: 
\begin{equation}
h_t^j = ( 1- z_t^j ) h_{t-1}^j + z_t^j \tilde{h}_t^j
\end{equation}
Where \(\tilde{h}_t^j\) and \(h_{t-1}^j\) correspond to the new candidate and previous memory content, respectively. \(z_t^j\) represents the update gate that enables the model to decide the amount of past information (from previous time steps) to be transferred along to the future and the amount of new memory content to be added.\\
To calculate the update gate \(z_t\) for time step t, we use the previous hidden states \(h_{t-1}\) and the current input \(x_t\) in the following equation:
\begin{equation}
z_t = \sigma(W_z x_t + U_z h_{t-1})
\end{equation}
The new memory content \(\tilde{h}_t^j\) is calculated as follows:
\begin{equation}
\tilde{h}_t = \tanh(W x_t + r_t \odot Uh_{t-1})
\end{equation}
where $\odot$ is the Hadamard product (also known as the element-wise product)
and \(r_t\) represent the reset gate which used to determine the amount of information to forget from the past. We use this formula to calculate it:
\begin{equation}
r_t = \sigma(W_r x_t + U_r h_{t-1})
\end{equation}
A graphical representation of the GRU unit is illustrated in \autoref{fig:Fig2}. \\
GRU is faster than LSTM on training since GRU has simpliﬁed architecture with fewer parameters and therefore uses less memory.

\subsubsection{Convolutional neural network (CNN)}
CNNs are generally used in computer vision, but lately, they have been extended to various NLP tasks and achieved impressive results in several languages and specifically in Arabic.
CNNs were added as a preliminary layer in our model used for the OTE task, to gain character-level features like word prefixes and suffixes \citep{chiu2016named}. We used CNN for training character vectors. The character vectors are then searched through the character lookup table to form the matrix C by stacking the searched results. Then various convolution filters of different sizes between the matrix C and multiple filter matrices are carried out to obtain the character-level features of each word by maximizing the pooling. Before inserting character embedding in CNN, we applied the dropout layer, which is useful for preventing the model from being dependent on certain words, this is an effective way to prevent overfitting.

\subsubsection{Bidirectional GRU (BGRU)}

One downside of GRU networks is that they can only use the previous context, without taking into consideration the future context, so they can only manage sequences from front to back which results in information loss. So many researchers have made use of bidirectional GRU that is capable of processing data in both directions, and information from both separate hidden layers is collected in the output layer. \\
The fundamental architecture of Bidirectional GRU networks is actually just putting together two separate GRUs. The input sequence is supplied for one network in normal time order (from right to left, for Arabic) and for another, in reverse time order. At each step, the outputs of both networks are usually combined (there are other options, such as summation). Such a structure can provide complete context information.

\subsubsection{Conditional random field (CRF)}

In this paper, the OTE task is considered as a sequence labeling task where, for a given aspect consisting of more than one word (in N-gram representation), It is useful to consider the associations between adjacent labels and decode them together by choosing their best sequence. We used the (inside, outside, beginning) IOB format to represent the N-gram aspects, where each token at the beginning of the aspect is marked as B-Aspect, I-Aspect if the token is located inside the aspect, not the first token, and O otherwise. In order to capture these correlations between aspect OTE tags, we applied conditional random field classifier (CRF) on top of BGRU. CRF is a standard model for predicting the most likely sequence of labels that correspond to a sequence of inputs. The CRF layer contains a state transition matrix to predict the current tag depending on both pre and post-tags. We refer to this transition matrix by Xi,j which represents the degree of transition from the i-th tag to the j-th tag. Then softmax is computed using the sentence scores and possible sequence of labels.The mathematical model was illustrated by the authors of \citep{lample2016neural}.

\subsection{Models}
In this part, we explain the proposed models in details: IAN-BGRU for polarity classification and  BGRU-CNN-CRF for aspect term extraction.

\subsubsection {IAN-BGRU for polarity classification}
\begin{figure}[ht!]
\begin{center}
 \includegraphics[width=.7\textwidth]{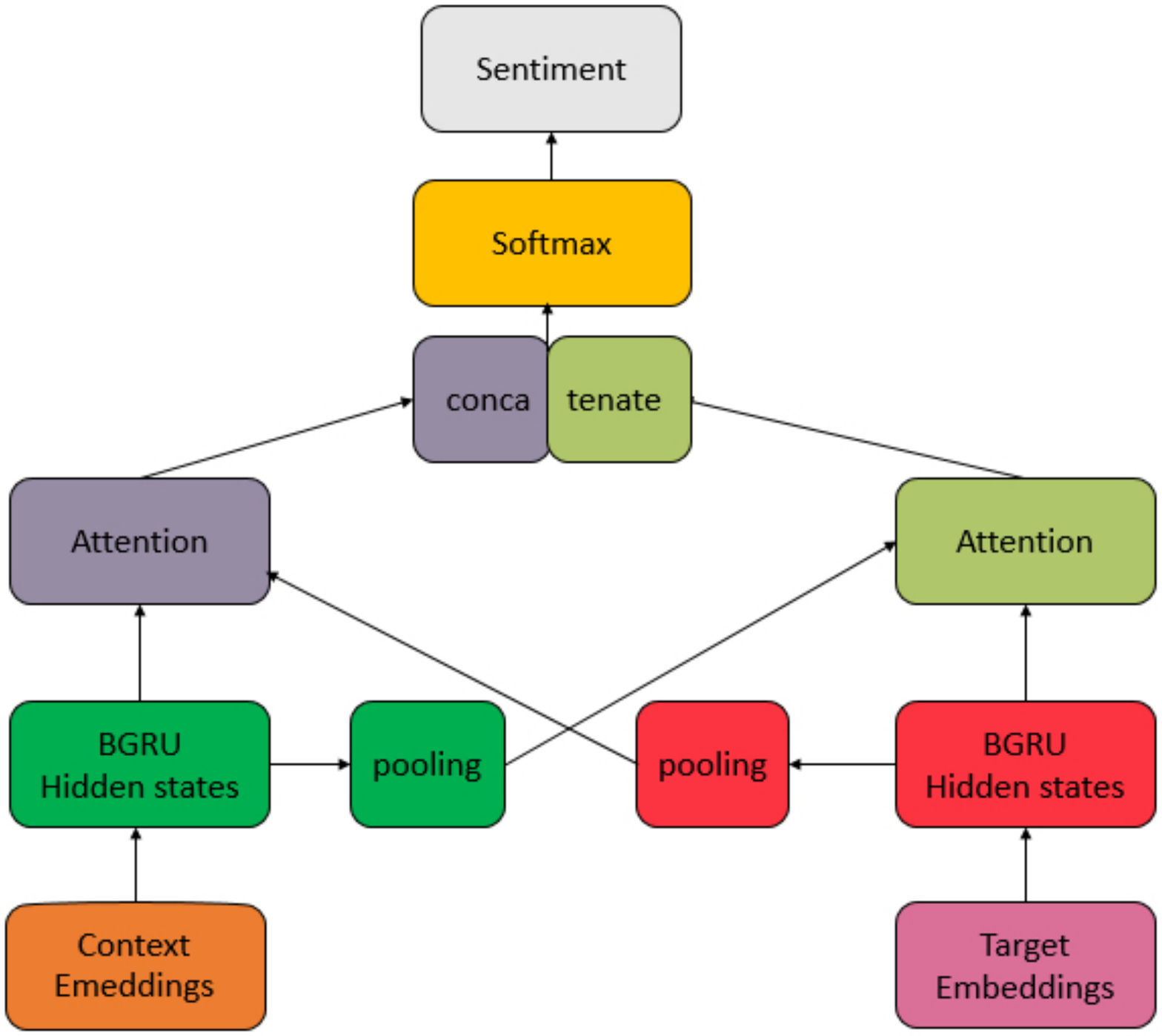}
    \caption{Architecture of the proposed IAN-BGRU model.}
    \label{fig:Figure3}
\end{center}
\end{figure}

To handle the third task, we applied the interactive attention network (IAN) \citep{ma2017interactive} with some modifications to the preliminary layers as IAN can precisely represent target and context interactively. This model consists of two parts: the first part considers obtaining the hidden states of the targets and their context by applying two separate LSTMs using word embeddings of target and context as input. To further improve the representation of targets and context we proposed using the features provided by GRU instead of LSTM in general and bidirectional GRU in particular. The bi-directional GRU resolves the feed-forward model's limited ability by extracting limitless contextual information from the front and back. Then, the hidden state values for both aspects and context are averaged separately, giving an initial representation of both, which will be used later to calculate the vectors of attention. The second part considers collecting essential data in the context through the application of the attention mechanism with the initial representation of the target leading to the context attention vector. Likewise, applying the attention mechanism with the initial representation of the context to capture important information in the target obtaining target attention vector. After that, the target and context representations can be defined based on the attention vectors. Ultimately, both target and context representations are concatenated and fed into a softmax function to identify sentiments for each aspect. Please refer to \citep{ma2017interactive} for basic model details.The overall architecture of the proposed IAN-BGRU model is depicted in \autoref{fig:Figure3}.

\subsubsection{BGRU-CNN-CRF for aspect term extraction}

\begin{figure}[ht!]
\begin{center}
 \includegraphics[width=.45\textwidth]{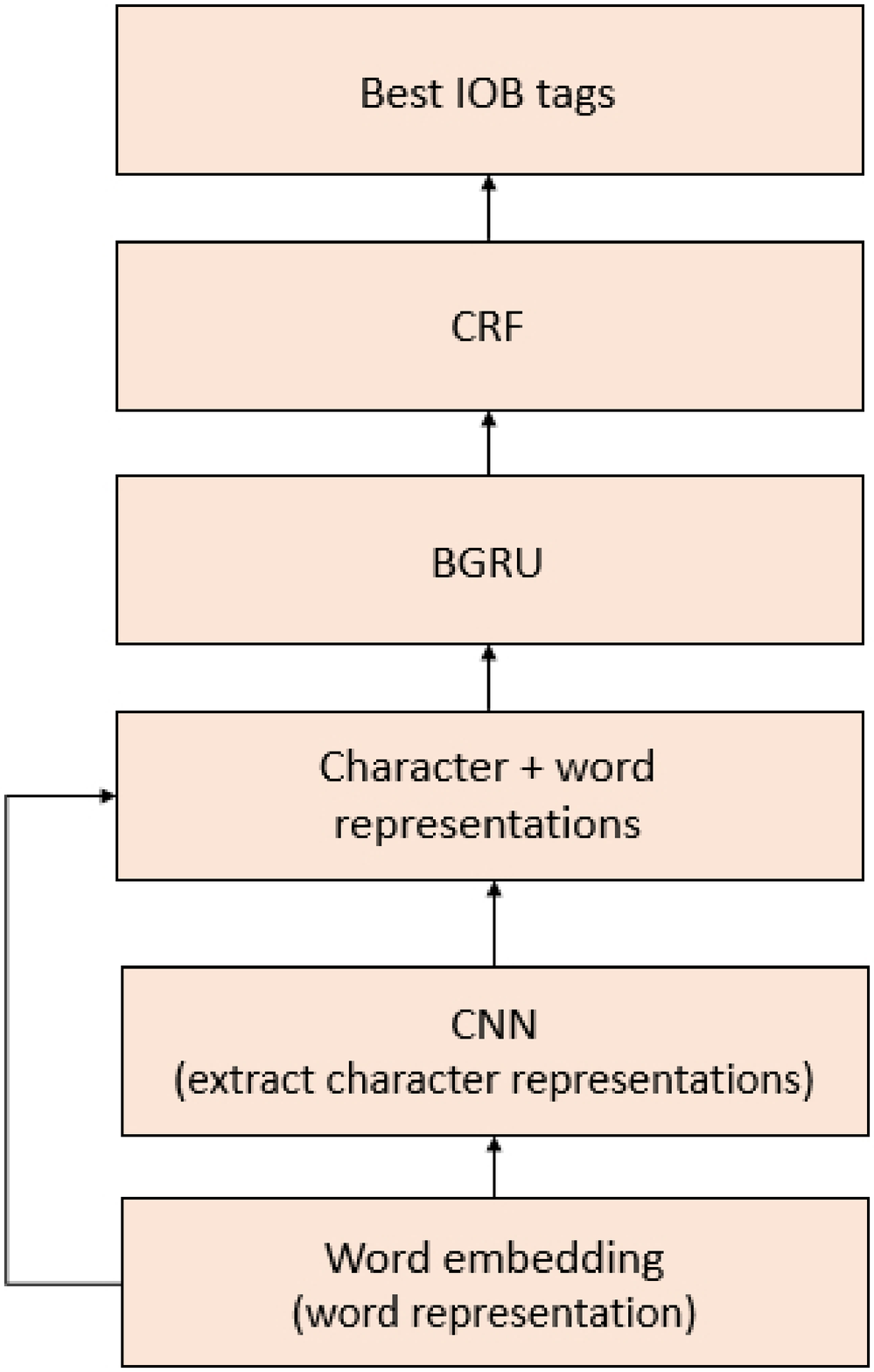}
    \caption{Architecture of the proposed BGRU-CNN-CRF model.}
    \label{fig:Figure4}
\end{center}
\end{figure}

In this paper, the OTE task was considered as a sequence labeling task, so we suggested a sequence labeling DL-based model proposed by the  authors \citep{ma2016end} for this task, where their model made use of character level representations and used the CRF layer as the output layer. Except for pre-trained word embedding in unlabeled corpora, it is an end-to-end model that does not rely on any handcrafted features or external resources.\\
First, we used CNN to encode word information at the character level to obtain a character-level representation. A dropout layer \citep{srivastava2014dropout} is applied before feeding the CNN with character embeddings. Then, character embedding vectors (character-level-representations, extracted previously by  CNN ) are concatenated with word embedding vectors and fed into the BGRU to model context information of each word. The dropout layer is also applied to output vectors from the BGRU. The output vectors of BGRU are then fed into the CRF layer for decoding the best sequence of labels. The overall architecture of the proposed BGRU-CNN-CRF model is depicted in \autoref{fig:Figure4}.

\section{Data and baseline research}

The main tasks of our research were tested using the Arabic hotel reviews dataset \footnote{ \url{ https://github.com/msmadi/ABSA-Hotels }}. The dataset was prepared as a part of SemEval-2016 Task-5 which was a multilingual ABSA task that includes customer reviews in eight languages and seven domains \citep{pontiki-etal-2016-semeval}. The Arabic hotel reviews dataset contains 24,028 annotated ABSA tuples divided as follow: 19,226 tuples for training and 4,802 tuple for testing. Furthermore, both text level (2291 reviews texts) and sentence level (6029 sentences) annotations were provided for the dataset. This research concentrated only on the tasks at sentence level. \autoref{tab:Tabel1}, indicates the size and distribution of the dataset to the tasks of the research.\\
The dataset was supplemented with baseline research based on SVM and N-grams as features. The results obtained from that research are considered as a baseline for each task and are mentioned in this paper in the results section related to each task.

\begin{table}[!ht]
\caption{The size and distribution of the dataset to the tasks of the research \citep{pontiki-etal-2016-semeval}.}
\begin{center}
\footnotesize
\begin{tabular}{ccccccccc}
    \hline
    \multicolumn{3}{c}{TASK}&\multicolumn{3}{c}{TRAIN}&\multicolumn{3}{c}{TEST}\\
    & & &text&sentence&tuples&text&sentence&tuples\\
    \hline
    \multicolumn{3}{c}{T1: Sentence-level ABSA}&1839&4802&10.509&452&1227&2604\\
    \hline
    \multicolumn{3}{c}{T2: Text-level ABSA}&1839&4802&8757&452&1227&2158\\
    \hline
\end{tabular}
\end{center}
\label{tab:Tabel1}
\end{table}

\section{Experimentation and results}
Both models (BGRU-CNN-CRF for T2, and IAN-BGRU for T3) were trained and tested using the Arabic hotel reviews dataset. For model training 70\% of the dataset was used, for validation 10\% was used and for testing 20\%. The Pytorch library was used to implement all the neural networks. The computations for each model were performed separately on the GeForce GTX 1080 Ti GPU. This section explains training of each model based on the targeted task.

\subsection {opinion target expression extraction task (T2)}
To handle the OTE task, BGRU-CNN-CRF model is implemented and trained using the features of word and character embedding.

\subsubsection{Evaluation method}

F1 metric is adopted for performance evaluation of OTE which is referred as the weighted harmonic mean for accuracy and recall. The score is computed by
\begin{equation}
F_1  = (\frac{recall^{-1} + precision ^{-1}}{2})^{-1} = 2 \cdot \frac{precision  \cdot recall}{precision  + recall}
\end{equation}

\subsubsection{Hyperparameters Setting}

Two types of word embeddings were proposed: a) fastText embeddings based on CBOW model. b) Word2Vec embeedings.\\
For Word2eVec: we intilaized both context word embeddings and targets word embeddings with AraVec \citep{soliman2017aravec} which is a pre-trained distributed word representation (word embedding) that intends to provide free-to-use and effective word embedding models for Arabic NLP research community. It is basically a word2vec model that has been trained on Arabic data. Two different models, unigrams and n-grams are provided by AraVec, which are built on top of various domains with Arabic content. In this research, we used CBOW-unigram model built on top of Twitter tweets,\\
We set the dimensions of word embeddings (for both fastText and Word2vec) and GRU hidden states to 100.\\
All character embeddings are initialized with uniform samples of\\ $[-\surd{3dim},+\surd{3dim}]$, where dim=25 in addition, 30 filters with a window size of 3 are used.\\
Mini-batch stochastic gradient descent (SGD) with batch size 16 and momentum 0.9 is used to optimize parameters. We selected an initial rate of ($\eta 0$=0.01), and at each training epoch, the learning rate is updated as $\eta t$=$\eta 0$/(1 +$\rho t$), with rate of decay $\rho$= 0.04, and t is the number of epochs completed. Gradient clipping of 5.0 was used to reduce the impact of “gradient exploding”.\\
The initial words embeddings are modified by updating the gradients of the neural network model via back-propagation. Early stopping \citep{caruana2001overfitting} was used based on validation set performance. The best parameters appear in about 60 epochs, according to our experiments.\\
In order to minimize overfitting, our model is regularized using the drop-out method.
Dropout is applied in several places in our model, namely : to character embeddings on all input and output vectors of GRU and before input to CNN. We fixed the dropout rate at 0.5 for all dropout layers.

\subsubsection{Results }

We did not find any published research on the same task that applied neural network models to the Arabic hotel reviews dataset and reported better results than \citep{al2019using} and \citep{al2020extracting}, so we compared our results with their.\\
The Authors of \citep{al2019using} have expanded the basic model BLSTM-CRF utilized for sequence labeling with character-level word embeddings by applying BLSTM to characters sequence in every word (BLSTM-CRF+LSTM-char). To initialize the word embedding lookup table, two techniques have been introduced.: (a) word2vec, (b) fastText. BLSTM-CRF+LSTM-char with word2vec word embeddings attained (F-1 = 66.32\%) and (F-1 = 69.98\%) with FastText embeddings.\\
The authors of \citep{al2020extracting} proposed attention based neural network to extract opinion targets. Their model composed of Bidirectional lstm with CNN as encoder and Bidirectional lstm as decoder with attention and CRF. Their model achieved F1-score = 72.83.

The differences between our model (CNN-BGRU-CRF) and the Bi-LSTM-CRF \citep{al2019using} are:
\begin{itemize}
    \item The BLSTM layer they used for character embedding has been substituted by the CNN layer. As CNN has fewer training parameters than BLSTM, training performance is higher and was recommended as the preferred approach as mentioned in \citep{zhai2018comparing} when compared the performance of BLSTM-CRF models with CNN-based and LSTM-based character-level embeddings for biomedical named entity recognition.
    \item We replaced the LSTM in the BLSTM-CRF base model with GRU due to the simplicity of the GRU (since GRUs have fewer parameters), ease of training, and thus ease of learning.
\end{itemize}

The differences between our model and the Attention-Based Neural Model \citep{al2020extracting} are:
\begin{itemize}
\item
    Our model (BGRU-CNN-CRF) relies mainly on Bi-GRU to extract word level features instead of Bi-LSTM.
\item
    Also, two different types of word embeddings were used to train our model (fastText and Word2vec).
\end{itemize}

\autoref{tab:Tabel2} provides a comparison of F1 scores between the proposed model, the baseline, and the above-mentioned models using the Arabic hotel reviews dataset for evaluation on T2.

\begin{table}[ht]
\caption{Comparison of F1 scores between the proposed model, the baseline, and other related works on T2.}
\begin{center}
 \begin{tabular}{c  c} 
 \hline
 Model & F1(\%) \\ [0.5ex] 
 \hline
 Baseline & 30.97 \\ 
 Bi-LSTM-CRF (word2vec) & 66.32 \\
 Bi-LSTM-CRF (fastText) & 69.98 \\
 CNN-BGRU-CRF (word2vec) & 69.44 \\
 CNN-BGRU-CRF (fastText) & 70.67 \\
 \textbf{Attention-Based Neural Model} & \textbf{72.83} \\
 \hline
\end{tabular}
\end{center}
\label{tab:Tabel2}
\end{table}

\begin{figure}[ht!]
\begin{center}
 \includegraphics[width=.6\textwidth]{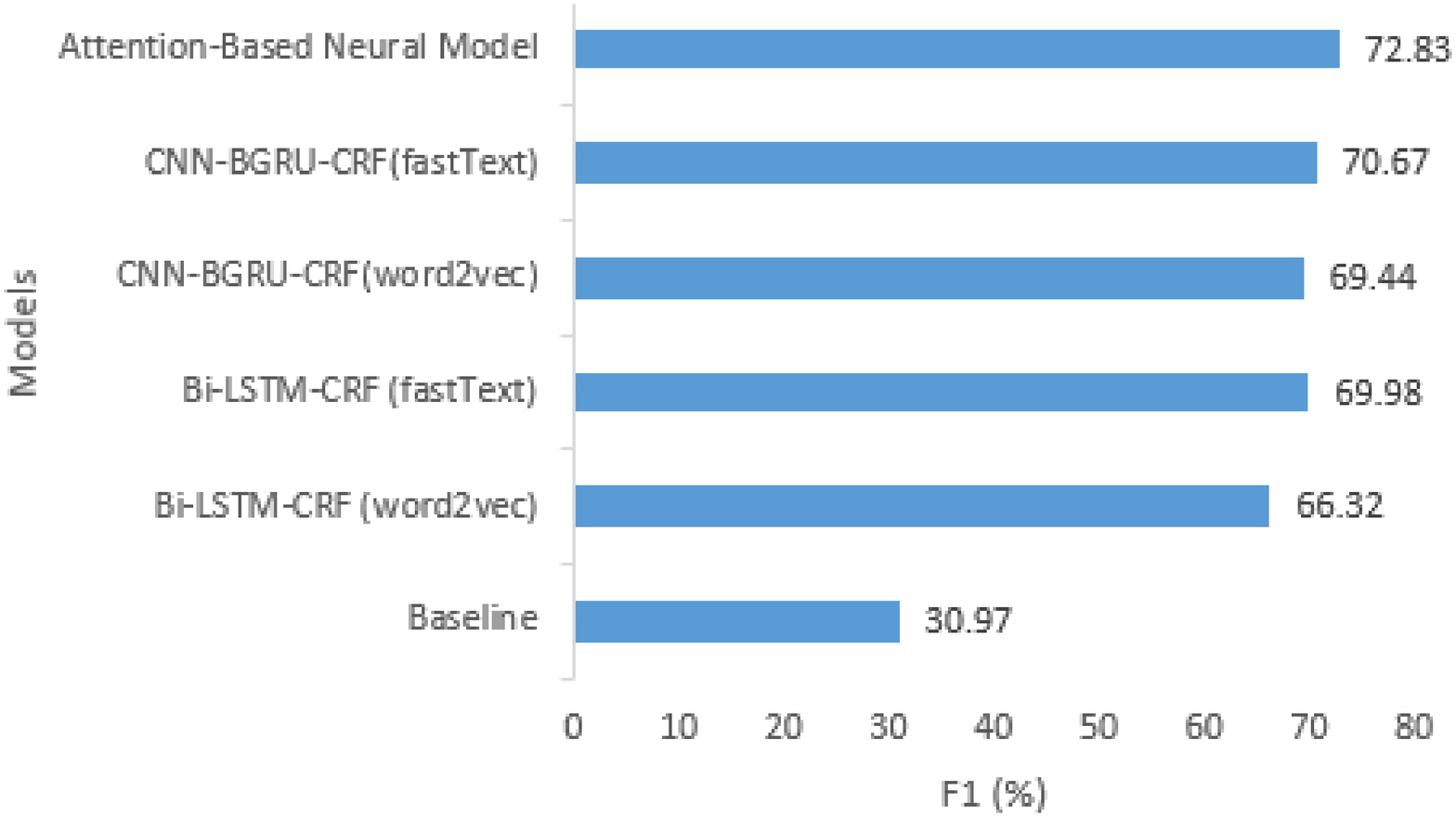}
    \caption{Achieved results in T2 (OTE) for the two proposed models (CNN-BGRU-CRF (word2vec) and CNN-BGRU-CRF (fastText)) compared to the baseline, and other related works.}
    \label{fig:Figure5}
\end{center}
\end{figure}

\autoref{fig:Figure5} shows that, despite the simplicity of our model, it outperformed the Bi-LSTM-CRF and achieved high results close to \citep{al2020extracting} when using the fastText embeddings.

\subsection{aspect sentiment polarity identification (T3)}
\subsubsection{Evaluation method}
To determine models efficiency in the aspect sentiment polarity identification task, accuracy metric is adopted that can be expressed as:

\begin{equation}
Accuracy = \frac{T}{N}
\end{equation}

where T and N, respectively, refer to the correctly predicted number of samples and  the overall samples number. Accuracy is the percentage of samples predicted correctly out of all samples. A better-performed system has better accuracy. 

\subsubsection{Hyperparameters Setting}

Both context word embeddings and targets word embeddings are initialized by AraVec \citep{soliman2017aravec}. In particular, the CBOW-unigram model built on top of Twitter tweets.\\
We initialize all weight matrices using samples from the uniform distribution  $U(-0.1,0.1)$.\\
we set the dimensions of word embeddings, attention vectors, and hidden states to 300.\\
After tunning the parameters of the model by making several runs using different parameters with specified transformations, we could found the combination of parameters that yield the best results, which are as follow: Back-propagation algorithm is used to train the model in sentence level using Adam \citep{kingma2014adam} for optimization with a learning-rate of $3e^{-5}$, L2-regularization weight of $2e^{-5}$, dropout rate of 0.3, batch size of 64 and number of epochs equal to 12.

\subsubsection{Comparison Models}
\noindent
\textbf{Baseline} *  SVM is trained only with N-grams features.\\
\textbf{INSIGHT-1} *  Won 1st place in the SemEval-2016 Task-5 competition evaluated on the Arabic hotel reviews dataset. They concatenated aspect embedding with every word embedding and fed the mixture to CNN  for aspect sentiment and category identification tasks \citep{ruder2016insight}.\\
\textbf{LSTM}  uses only one LSTM to model the sentence and the last hidden states as a representation for final classification.\\
\textbf{TD-LSTM} splits the sentence into the left and right parts around the aspect and flows into two LSTM models in different forward and backward sequential paths. In order to determine the sentiment polarity label, the final hidden vectors of the left LSTM and the right LSTM are linked to be fed into Softmax \citep{tang2015effective}.\\
\textbf{AB-LSTM-PC} *  attention-based LSTM with Aspect integration uses the attention mechanism that allows to focus more on the context relevant to the targeted aspects. For each word embedding, ATAE-LSTM adds aspects embedding that reinforce the model by learning the hidden association between context and aspect \citep{wang2016attention}. ATAE-LSTM model was applied by the authors of \citep{al2019using} (AB-LSTM-PC) on Arabic hotel reviews dataset.\\
\textbf{MemNet}  in order to correctly catch the importance of the contextual words, MemNet applied multi-hop attention layers on context word embeddings of the sentence and considered the last hop output as the final target representation \citep{tang2016aspect}.\\
\textbf{IAN-LSTM}  employs two LSTMs for interactive modeling of context and target. Context hidden states are used to generate the target attention vector, and target hidden states are used to generate the context attention vector. On the basis of these two attention vectors, context representations and target representation are created, then concatenated and ultimately fed into softmax for classification \citep{ma2017interactive}.\\
\textbf{IAN-BLSTM}  extends IAN-LSTM by using bidirectional LSTM instead of uni-direction LSTM to model aspect term and the context. \\
\textbf{IAN-GRU}  like IAN-LSTM but use GRU instead of LSTM to model aspect term and its context.\\
\textbf{IAN-BGRU}  extends IAN-GRU by using bidirectional GRU instead of uni-direction GRU to model the aspect term and context. \\
Models with * sign mean that, results were adopted from their research  without being practically re-implemented. \\
To our knowledge, no other research has applied models without * sign on Arabic ABSA in general and on Arabic hotel reviews in particular. 
\autoref{tab:Tabel3} provides a comparison of the accuracy results between the proposed model and the above-mentioned models using the Arabic hotel reviews dataset for evaluation on T3.

\begin{table}[ht]
\caption{ Comparison of accuracy results between the proposed model, the baseline, and other Arabic models on T3.}
\begin{center}
 \begin{tabular}{c c} 
 \hline
 Model & Accuracy \\ [0.5ex] 
 \hline
 Baseline & 76.4 \\ 
 INSIGHT-1 (CNN)* & 82.71 \\
 LSTM & 81.49 \\
 TD-LSTM & 81.79 \\
 AB-LSTM-PC* & 82.60 \\
 MEMNET & 82.64 \\
 IAN-LSTM & 83.18 \\
 IAN-GRU & 83.68 \\
 IAN-BLSTM & 83.48 \\
 \textbf{IAN-BGRU} & \textbf{83.98}\\
 \hline
\end{tabular}
\end{center}
\label{tab:Tabel3}
\end{table}

\subsubsection{Discussions}

\begin{figure}[ht!]
\begin{center}
 \includegraphics[width=.6\textwidth]{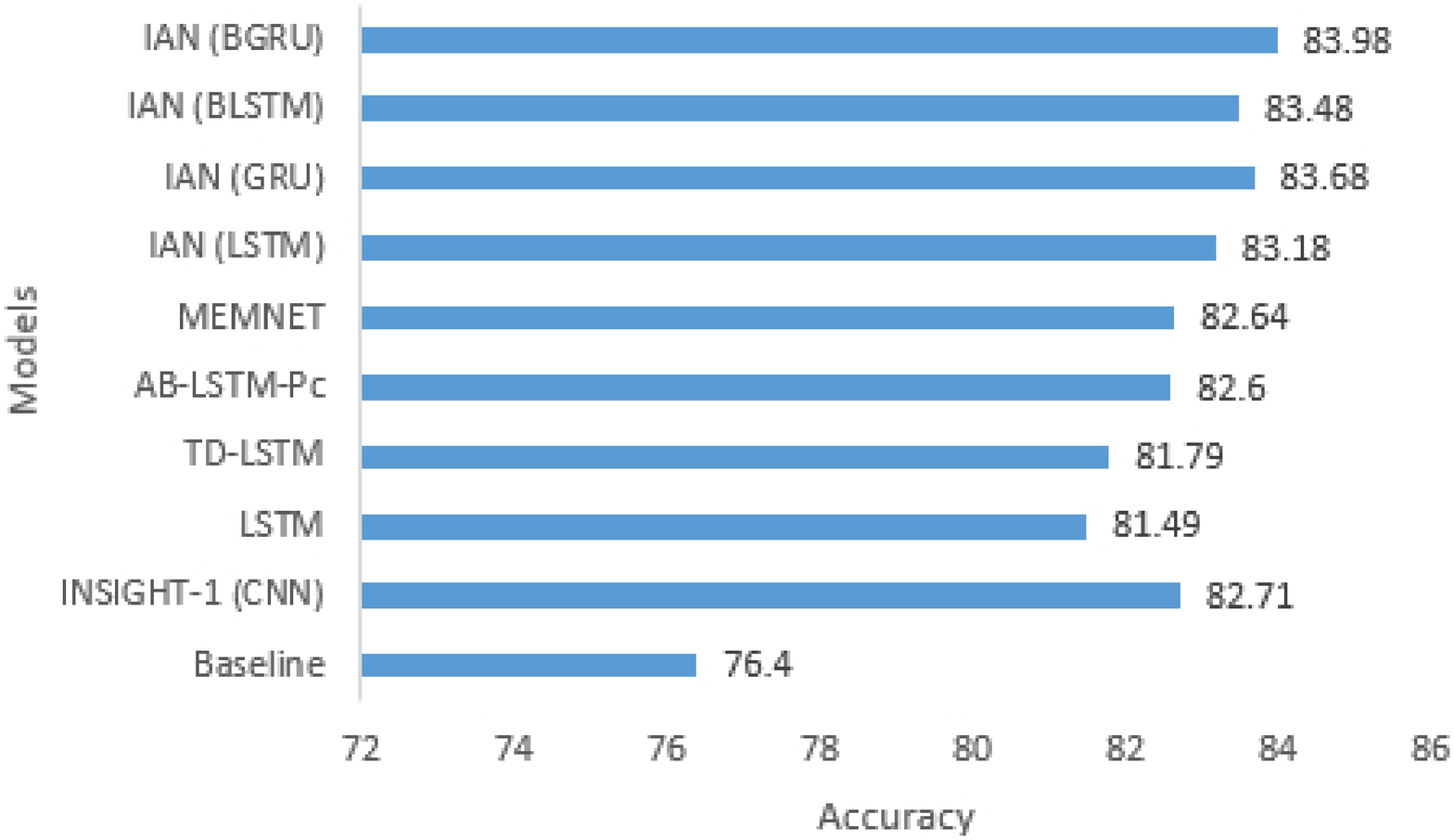}
    \caption{Achieved results by the IAN-BGRU model in comparison to the baseline and other Arabic models on T3.}
    \label{fig:Figure6}
\end{center}
\end{figure}

LSTM achieves the poorest performance out of all neural network baseline methods because it deals with targets on a par with other context words. As the target information is not used sufficiently.\\
TD-LSTM outperforms LSTM, as it evolves from the standard LSTM and handles both left and right contexts of the targeted aspect separately. The targets are represented twice and are emphasized in certain ways in the final representation. \\
Moreover, AB-LSTM-PC stably outperforms TD-LSTM for its introduction of the attention mechanism. As it collects a range of significant contextual information, under the guidance of the target and creates more accurate representations for ABSA.
AB-LSTM-PC also affirms the importance of modeling targets by including embeddings of aspect, which is the cause for enhancement in results. \\
INSIGHT-1(CNN) adopted the same idea by emphasizing the modeling of targets by concatenating the aspect embedding with every word embedding and applied convolution layer over it, trying to simulate the effect of attention by using a set of filters that could capture the features in a better manner. That is the reason for performance improvement over ATAE-LSTM.\\
IAN models emphasize the significance of targets via interactive learning of targets and context representations. We can see that across all baselines, IAN provides the best performance. The key explanation for this could be that IAN relies on two related attention networks that affect one another for modeling target and context interactively. IAN(BLSTM) outperform IAN(LSTM) since it could resolve the feed-forward model's limited ability by extracting limitless contextual information from the front and back to model aspect term and the context. We noticed that IAN(GRU) model achieved better results than both IAN(LSTM) and IAN(BLSTM) by about 0.5\% and 0.25\% when replacing LSTM by GRU, that is due to the simplicity of GRU (since GRUs have less number of parameters) and ease of training and consequently ease of learning. And by using (IAN-BGRU) bidirectional GRU instead of unidirectional, we were able to obtain the best results among all the methods as shown in \autoref{fig:Figure6}.

After finding a noticeable improvement in performance when using models that apply attention mechanism, we tried to adopt the idea of multiple attentions to synthesize important features in difficult sentence structures using models such as MEMNET. However, we noticed no improvement in performance. When using MEMNET, we found a slight improvement in performance compared to the AB-LSTM-PC model and lower results than all IAN variations, that is maybe due to the complexity of the model and need a mechanism to stop the attention process automatically if no more useful information can be read. 

\subsubsection{Case study}

The importance of IAN-BGRU comes from relying on two related attention networks that affect one another for interactive modeling of context and target. The model can closely focus on important parts of the target and context and ultimately provide a clear representation of both, that enabled the model to classify the different sentiments that refer to different aspects of the same sentence.\\
\noindent
As shown in this review sentence : \\
\<“موقع جيدا جدا و وسائل ترفيه متوفرة لكن الغرف كانت للاسف سيئة” ”>\\
in English "very good location and means of entertainment available but the rooms were unfortunately bad". This review has two different aspects (location and rooms) with two different sentiments (pos and neg). Even with this difficulty, our model can correctly recognize and determine the expressed polarity of each aspect.\\
The main challenge of sentiment analysis is the ability of classifying the reviews sentiment polarity if there are shifting words like “not”. This is due to the issue that, it completely alters the polarity of aspects. For example, in this sentence “The hotel location is not good for the elderly”.Traditional SA methods need shifting words lexicons for classifying  the aspect polarity (Hotel\#Location) as Negative. But our model can figure this out without needing any external source like lexicons.

\section{Conclusion and future work}
ABSA provides us with more detailed information than SA, as it can define the aspects of a given document or sentence and the sentiments conveyed regarding each aspect. Three main ABSA tasks can be identified : T1: aspect category identification, T2: aspect opinion target extraction, and T3: aspect polarity detection. The topics of this study are T2 and T3. Two GRU based models were adopted to handle research work. (a) DL model that takes advantage of word and character representations by combining bidirectional GRU, Convolutional Neural Network (CNN), and Conditional Random Field (CRF) making up the (BGRU-CNN-CRF) model to extract the main opinionated aspects (OTE). (b) an interactive attention network based on bidirectional GRU (IAN-BGRU) to identify sentiment polarity toward extracted aspects. We evaluated our models using the benchmarked Arabic hotel reviews dataset. 
The results indicate that the proposed methods are better than baseline research on both tasks having 39.7\% enhancement in F1-score for opinion target extraction (T2) and 7.58\% in accuracy for aspect-based sentiment polarity classification (T3). Achieving F1 score of 70.67\% for T2, and accuracy of 83.98\% for T3.\\
For future work, we intend to apply transformer based models in ABSA tasks.

\bibliography{bibliography}
\typeout{get arXiv to do 4 passes: Label(s) may have changed. Rerun}
\end{document}